\documentclass[10pt,twocolumn,letterpaper]{article}
\usepackage{iccv}
\usepackage{times}
\usepackage{epsfig}
\usepackage{graphicx}
\usepackage{amsmath}
\usepackage{amssymb}
\usepackage[numbers]{natbib} 
\usepackage[pagebackref=true,breaklinks=true,letterpaper=true,colorlinks,bookmarks=false]{hyperref}
\iccvfinalcopy 
\def\iccvPaperID{10428} 
\ificcvfinal\fi
\makeatletter
\@namedef{ver@everyshi.sty}{}
\makeatother

\usepackage{amsmath,amsfonts,bm}
\usepackage{algorithm,algorithmic}

\def\eqref#1{equation~\ref{#1}}

\def\1{\bm{1}}

\def\vomega{{\bm{\omega}}}

\def\vm{{\bm{m}}}

\def\vx{{\bm{x}}}
\def\vy{{\bm{y}}}
\def\vz{{\bm{z}}}

\DeclareMathAlphabet{\mathsfit}{\encodingdefault}{\sfdefault}{m}{sl}
\SetMathAlphabet{\mathsfit}{bold}{\encodingdefault}{\sfdefault}{bx}{n}

\newcommand{\R}{\mathbb{R}}

\DeclareMathOperator*{\argmax}{arg\,max}

\usepackage{amssymb}

\usepackage[inline, shortlabels]{enumitem}  \usepackage{nicefrac}       \usepackage{booktabs}       \usepackage{multirow}
\usepackage{colortbl}
\usepackage[export]{adjustbox}
\usepackage{float}
\usepackage{arydshln}
\usepackage{tikz}
\usetikzlibrary{arrows,positioning,calc}
\usepackage{dblfloatfix}
\usepackage{placeins} 
\begin{document}

\title{Semantic Perturbations with Normalizing Flows for Improved Generalization}

\author{Oğuz Kaan Yüksel$^\dag$
\qquad Sebastian U. Stich$^\dag$
\qquad Martin Jaggi$^\dag$
\qquad Tatjana Chavdarova$^{\dag, \ddagger}$\\
$^\dag$ Machine Learning and Optimization Lab, EPFL\\
$^\ddagger$ Department of Electrical Engineering and Computer Sciences, UC Berkeley\\
\\}

\maketitle
\ificcvfinal\thispagestyle{empty}\fi

{\let\thefootnote\relax\footnotetext{Correspondence to {\tt oguz.yuksel@epfl.ch}.}}
\begin{abstract}
Data augmentation is a widely adopted technique for avoiding overfitting when training deep neural networks.
However, this approach requires domain-specisfic knowledge and is often limited to a fixed set of hard-coded transformations.
Recently, several works proposed to use generative models for generating semantically meaningful perturbations to train a classifier. However, because accurate encoding and decoding are critical, these methods, which use architectures that approximate the latent-variable inference, remained limited to pilot studies on small datasets. 

Exploiting the exactly reversible encoder-decoder structure of normalizing flows, we perform on-manifold perturbations in the latent space to define fully unsupervised data augmentations.
We demonstrate that such perturbations match the performance of advanced data augmentation techniques---reaching $96.6\%$ test accuracy for CIFAR-10 using ResNet-18 and outperform existing methods, particularly in low data regimes---yielding $10$--$25\%$ relative improvement of test accuracy from classical training. We find that our latent \emph{adversarial} perturbations adaptive to the classifier throughout its training are most effective, yielding the first test accuracy improvement results on real-world datasets---CIFAR-10/100---via latent-space perturbations. 
\end{abstract}

\section{Introduction}
Deep Neural Networks (DNNs) have shown impressive results across several machine learning tasks~\cite{resnet,mnih2015humanlevel}, and---due to their automatic feature learning---have revolutionized the field of computer vision. However, their success depends on the availability of large annotated datasets for the task at hand. Thus, among other overfitting techniques---such as L1/L2 regularization, dropout~\cite{srivastava2014dropout}, early stopping, among others---data augmentation remains a mandatory component that is frequently used in practice. 

Traditional data augmentation (DA) techniques apply a predefined set of transformations to the training samples that do not change the corresponding class label, to increase the number of training samples. As this approach is limited to making the classifier robust only to the fixed set of hard-coded transformations, advanced methods incorporate more loosely defined transformations in the data space. For example, \textit{mixup}~\cite{zhang2018mixup} uses convex combinations of pairs of examples and their labels, and \textit{cutout}~\cite{devries2017cutout} randomly masks square regions of the input sample. Albeit implicitly, these methods still require domain-specific knowledge that, for example, such masking will not change the label.

Surprisingly, in the context of computer vision, it has been shown that small perturbations in image space that are not visible to the human eye can fool a well-performing classifier into making wrong predictions. This observation motivated an active line of research on adversarial training~\cite[see][and references therein]{Biggio18adversarial}---namely, training with such adversarial samples to obtain robust classifiers. However, further empirical studies showed that such training reduces the training accuracy, indicating the two objectives are competing~\cite{tsipras2019robustness,su2018robustness}. 

\citet{stutz2019disentangling} postulate that this robustness-generalization trade-off appears due to using off-manifold adversarial attacks that leave the data-manifold and that \emph{on-manifold adversarial attacks} can improve generalization. 
For verifying this hypothesis, the authors proposed to use perturbations in the latent space of a generative model. Their proposed method employs (class-specific) models named VAE-GANs~\cite{larsen2016,rosca2017variational}---which are based on Generative Adversarial Networks~\cite{goodfellow2014generative} and, to tackle their non-invertibility, further combine GANs with Variational Autoencoders~\cite{kingma2014vae}.
However, the VAE-GAN model introduces hard-to-tune hyperparameters, and notably, it optimizes a lower bound on the log-likelihood of the data. Moreover, improved test accuracy was only shown on toy datasets~\cite[Fig. 5]{stutz2019disentangling}, and yet in some cases, the test accuracy did not improve relative to classical training. We observe that on real-world datasets, such training can decrease the test accuracy, see~\S\ref{sec:experiments}.

In this work, we focus on the possibility of employing advanced normalizing flows such as \textit{Glow}~\cite{kingma2018glow}, to define entirely unsupervised augmentations---contrasting with pre-defined fixed transformations---with the same goal of improving the generalization of deep classifiers. 
Although normalizing flows have gained little attention in our community relative to GANs and Autoregressive models, they offer appealing advantages over these models, namely:
\begin{enumerate*}[series = tobecont, itemjoin = \quad, label=(\roman*)]
\item exact latent-variable inference and log-likelihood evaluation, and 
\item efficient inference and synthesis that can be parallelized~\cite{kingma2018glow}, respectively.
\end{enumerate*}
We exploit the \emph{exactly reversible} encoder-decoder structure of normalizing flows to perform efficient and controllable augmentations in the learned manifold space.

\vspace{-0.2em}
\paragraph{Contributions.} Our contributions can be summarized as:\begin{itemize}[leftmargin=12pt,parsep=1pt,topsep=1pt]
    \item Firstly, we demonstrate through numerical experiments that the previously proposed methods to generate on-manifold perturbations fail to improve the generalization of a trained classifier on real-world datasets. In particular, the test accuracy decreases with such training on CIFRAR-10/100. In this work, we postulate that this occurs due to \emph{approximate} encoder-decoder mappings.
    \item Motivated by this observation, we propose a data augmentation method based on \emph{exactly reversible} normalizing flows. Namely, it first trains the generative model and then uses simplistic random or adversarial domain-agnostic semantic perturbations to train the classifier, defined in \S\ref{sec:latent_attacks}.
    \item We demonstrate that our adversarial data augmentation method generates on-manifold and semantically meaningful data perturbations. Hence, we argue that our technique is a novel approach for generating perceptually meaningful (natural adversarial examples), different from previous proposals.
    \item Finally, we empirically demonstrate that our on-manifold perturbations consistently outperform the standard training on CIFAR-10/100 using ResNet-18. Moreover, in a low-data regime, such training yields up to $25\%$ relative improvement from classical training, of which---as most effective---we find the adversarial perturbations that are adaptive to the classifier, see~\S\ref{sec:experiments}.
\end{itemize}

\section{Related Work}\label{sec:related_works}
\noindent\textbf{Data augmentation} techniques are routinely used to improve the generalization of classifiers~\cite{Simard1998,krizhevsky2017imagenet}. While most classic techniques require a priori expert knowledge of invariances in the dataset to generate \emph{virtual} examples in the vicinity around each sample in the training data, many automated techniques have been proposed recently, such as linearly interpolating between images and their labels \cite{zhang2018mixup}, replacing a part of the image with either a black-colored patch \cite{devries2017cutout} or a part of another image \cite{yun2019cutmix}.

In contrast to these data-agnostic procedures,
a few recent works proposed to learn useful data augmentations policies, for instance by optimization~\cite{Fawzi2016:adaptive,Ratner2017:transformations}, reinforcement learning techniques~\cite{cubuk2019autoaugment,Cubuk2020:rand,Zhang2019:advauto}, specifically trained augmentation networks~\cite{Peng2018:jointly,tang2020onlineaugment} or assisted by generative adversarial networks~\cite{Perez2017:dataaugment,Antoniou2017:augmentation,Zhang2018:metagan,Tran2020:gan}, such as also in \cite{mikolajczyk2019style} that proposes neural style transfer for augmenting datasets. 

\noindent\textbf{Perturbations in Latent Space}
allow natural data augmentation with GANs. For instance,
\citet{Antoniou2017:augmentation,Zhao2018:generating} propose to apply random perturbations in the latent space,  and recently \citet{manjunath2020improving} used StyleGAN2 \cite{karras2020analyzing} to generate novel views of the image through latent space manipulation.
However, a critical weakness in these techniques is that the mapping from the latent space to the training data space is typically not invertible, \textit{i.e.}, finding the representation of a data sample in the latent space (to start the search procedure) is a non-trivial task. For instance, 
\citet{Zhao2018:generating} propose to separately train an inverter for the inverse-mapping to the latent space. 
This critical bottleneck is omitted in our approach since we rely on an invertible architecture which renders the learning of an inverter superfluous.

\noindent\textbf{Latent attacks}, \textit{i.e.}, searching in the latent space to find virtual data samples that are misclassified, were proposed in \cite{Baluja2017:adversarialTN,Song2018:gan,Xiao2018:generating,Zhang2020:auxiliary}.
\citet{volpi2018generalizing} proposed an adaptive data augmentation method that appends adversarial examples at each iteration and note that generalization is improved across a range of a priori unknown target domains.
Complementary, the connection of adversarial learning and generalization has also been studied in~\cite{Tanay2016:manifold,Rozsa2016:robustness,Jalal2017:manifold,tsipras2018robustness,Gilmer2018:spheres,Zhao2018:generating}.
\citet{stutz2019disentangling} clarify the relationship between robustness and generalization by showing in particular that regular adversarial examples leave the data
manifold and that on-manifold adversarial training boosts generalization. These important insights endorse previous findings that data augmentation assisted by generative models---as we suggest here---can improve generalization~\cite{volpi2018generalizing}.

\noindent\textbf{Perceptual (or \emph{Natural}) Adversarial examples} are getting increasing interest in the community recently, as alternative to---from human perceptive---often hard to interpret standard adversarial threat models~\cite{Zhao2018:generating,Robey2020:model,wong2020learning,Luo2020:dataaugmentation,Laidlaw2021:peceptual,kireev2021effectiveness, dolatabadi2020advflow}.
We argue that on-manifold perturbations, as obtained with our method or similar generative techniques, can
 \emph{implicitly learn} such natural transformations and could be used as an alternative method to define and generate perceptually and semantically meaningful data augmentation.
 In contrast to \citet{wong2020learning} who
 propose to learn \emph{perturbation} sets via a latent space of a conditional variational autoencoder using a set of predefined image-space transformations, in our approach, we are not restricted to a fixed transformation set as we utilize implicit transformations learned by the invertible mapping provided by normalizing flows.

\section{Normalizing Flows and their Advantages for Semantic Perturbations}\label{sec:nf}

In this section, we first describe the fundamental concepts of normalizing flows.
We then discuss how their ability to perform exact inference helps to apply perturbations in latent space.

\subsection{Background: Normalizing Flows}

Assume observations $\vx \in \R^d$ sampled from an unknown data distribution $p_{\mathcal{X}}$ over $\mathcal{X} \subset \R^d$, 
and a tractable prior probability distribution $p_{\mathcal{Z}}$ over $\mathcal{Z} \subset \R^{k}$ according to which we sample a latent variable $\vz$.
Flow-based generative models seek to find an invertible, also called \emph{bijective} function $\mathcal{F}: \mathcal{X} \rightarrow \mathcal{Z}$ such that: 
\begin{equation}\tag{NF}\label{eq:nf}
    \mathcal{F}(\vx) = \vz \quad \text{and} \quad \mathcal{F}^{-1}(\vz) = \vx \,,
\end{equation}
with $\vz \in \mathcal{Z}$ and $ \vx \in \mathcal{X}$.
That is, $\mathcal{F}$ maps observations~$\vx$ to latent codes $\vz$, and $\mathcal{F}^{-1}$ maps latent codes $\vz$ back to original observations $\vx$.

The key idea behind normalizing flows is to use change of variables, \textit{i.e.}, by using invertible transformation we keep track of the change in distribution. Thus, $p_{\mathcal{X}}$ induces $p_{\mathcal{Z}}$ through~$\mathcal{F}$ and the opposite holds through $\mathcal{F}^{-1}$. We have:
\begin{align*}
    p_{\mathcal{X}}(\vx) &= p_{\mathcal{Z}}(\mathcal{F}(\vx)) \cdot \Big|\det \Big(\frac{\partial \mathcal{F}(\vx)}{\partial \vx^\top}\Big)\Big| \,, 
\end{align*}
where the determinant of the Jacobian $\frac{\partial \mathcal{F}(\vx)}{\partial \vx^\top}$ is used as volume correction.
In practice, $\mathcal{F}$ is also differentiable and is parameterized with parameters $\vomega$, we have finite samples $\vx_i\sim p_{\mathcal{D}}, 1\leq i\leq N$ and training is done via maximum log-likelihood:
\begin{equation*}
\resizebox{1\linewidth}{!}{$ \displaystyle \vomega^\star {=} \argmax_{\vomega} \sum_{i=1}^N \log p_{\mathcal{Z}}(\mathcal{F}(\vx_i|\vomega)) + \log \Big| \det \Big(\frac{\partial \mathcal{F}(\vx_i|\vomega)}{\partial \vx_i^\top}\Big)\Big| \,. $}
\end{equation*}
Because computing the inverse and the determinant is computationally expensive for high-dimensional spaces, $\mathcal{F}$ is constrained to linear transformations that have some structure---often chosen to be \emph{triangular} Jacobian matrices, which provide efficient computations in both directions. 

To build an expressive but tractable $\mathcal{F}$, we rely on the fact that differentiable functions are closed under composition, thus $\mathcal{F} = f_\ell \circ f_{\ell-1} \circ \cdots \circ f_1$, $\ell{>}1$, is also invertible. In the context of deep learning, this implies that we can stack $\ell$ layers of simple invertible mappings.
However,  as this still yields a single linear transformation, 
\textit{coupling} layers~\cite{dinh2017}  $f(\mathbf{x}) = \vy$, with $f:\R^C\rightarrow\R^C $ are inserted, which can be defined in several ways~\cite{Dinh2015nice,pmlr-v97-ho19a}. 
In this work we use affine coupling transforms, which are empirically shown to perform particularly well for images, and which are used in the \textit{Glow} model~\cite{kingma2018glow}: 
\begin{equation*}
\resizebox{1\linewidth}{!}{$ \displaystyle     \vy_{1:c} {=} \vx_{1:c} \quad \text{and}\quad 
    \vy_{c+1:C} {=} \vx_{c+1:C} \odot \exp(s(\vx_{1:c})) + t(\vx_{1:c})  \,,
    $}
\end{equation*}
where $\odot$ is the Hadamard product and, $s$ and $t$ are scaling and translations functions from $\R^c \rightarrow \R^{C-c}$.
Moreover, the Jacobian does not require any derivative over $s$ and $t$, meaning that we can model these functions with arbitrary deep neural networks.
To allow that each component can change, usually, $\mathcal{F}$ is composed so that coupling layers are placed in the middle of permutation layers that work in alternating patterns.
The \textit{Glow} model~\cite{kingma2018glow} uses an invertible 1$\times$1 convolution layer that generalizes this permutation operation, see Appendix~\ref{app:arch}.

\subsection{Advantages of Normalizing Flows}
\label{sec:advantages}

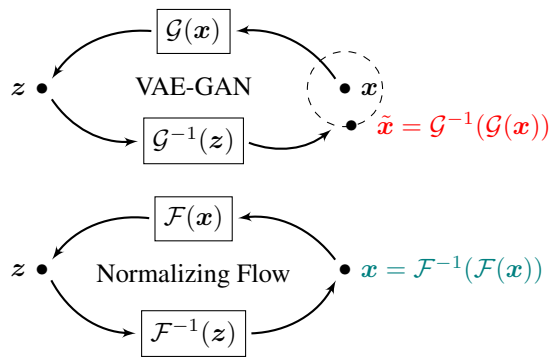
\begin{figure}[tb]
\centering
\begin{tikzpicture}[node distance=1cm, auto]
\tikzset{
myvar/.style={rectangle, text centered}, 
mymod/.style={rectangle, text centered, draw=black}, 
myarw/.style={->, >=latex', shorten >=2pt, shorten <=2pt,thick}, 
mylab/.style={}
}
\node(center)[myvar] at (0,0) {};
\node(VAEr)[above right=1cm and 1.8cm of center,inner sep=0pt, outer sep=0pt]{$\bullet$};
\node(NFr)[below right=1cm and 1.8cm of center,inner sep=0pt, outer sep=0pt]{$\bullet$};
\node(VAEl)[above left=1cm and 1.8cm of center,inner sep=0pt, outer sep=0pt]{$\bullet$};
\node(NFl)[below left=1cm and 1.8cm of center,inner sep=0pt, outer sep=0pt]{$\bullet$};
\node(VAE)[above=1cm of center,inner sep=0pt, outer sep=0pt]{VAE-GAN};
\node(NF)[below=1cm of center,inner sep=0pt, outer sep=0pt]{Normalizing Flow};
\draw[dashed] (VAEr) circle (0.5cm); 
\node(error) at ($(VAEr) + (280:0.5cm)$) {$\bullet$};
\node(VAEe)[mymod, above=0.3cm of VAE]{$\mathcal{G}(\vx)$};
\node(VAEd)[mymod, below=0.3cm of VAE]{$\mathcal{G}^{-1}(\vz)$};
\node(NFe)[mymod, above=0.3cm of NF]{$\mathcal{F}(\vx)$};
\node(NFd)[mymod, below=0.3cm of NF]{$\mathcal{F}^{-1}(\vz)$};
\draw[myarw,bend right=30] (VAEr.north west) to (VAEe.east);
\draw[myarw,bend right=30] (VAEe.west) to (VAEl.north east);
\draw[myarw,bend right=30] (NFr.north west) to (NFe.east);
\draw[myarw,bend right=30] (NFe.west) to (NFl.north east);
\draw[myarw,bend right=30] (NFl.south east) to (NFd.west);
\draw[myarw,bend right=30] (NFd.east) to (NFr.south west);
\draw[myarw,bend right=30] (VAEl.south east) to (VAEd.west);
\draw[myarw,bend right=30] (VAEd.east) to (error.west);
\node[auto, right=0cm of VAEr] {$\vx$};
\node[auto, right=0cm of NFr,teal] {$\vx = \mathcal{F}^{-1}(\mathcal{F}(\vx))$};
\node[auto, left=0cm of VAEl] {$\vz$};
\node[auto, left=0cm of NFl] {$\vz$};
\node[auto, right=0cm of error, red] {$\tilde \vx = \mathcal{G}^{-1}(\mathcal{G}(\vx))$};
\end{tikzpicture}
\caption{\textbf{Exactness of NF encoding-decoding}. Here~$\mathcal{F}$ denotes the bijective NF, and $\mathcal{G}/\mathcal{G}^{-1}$ encoder/decoder pair of inexact methods such as VAE or VAE-GAN which, due to inherent decoder noise, is only approximately bijective. }
\label{fig:schema_diff}
\end{figure}

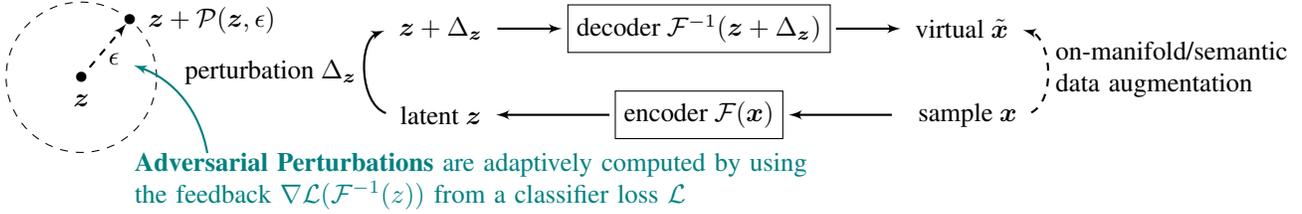
\begin{figure*}[t]  
\centering
\begin{tikzpicture}[node distance=1cm, auto]
\tikzset{
myvar/.style={rectangle, text centered, minimum height=18}, 
mymod/.style={rectangle, text centered, minimum height=18, draw=black}, 
myarw/.style={->, >=latex', shorten >=2pt, shorten <=2pt,thick}, 
mylab/.style={}
}
\node(generator)[mymod]{decoder $\mathcal{F}^{-1}(\vz+\Delta_\vz)$};
\node(inverter)[mymod, below=0.5cm of generator]{encoder $\mathcal{F}(\vx)$};
\node(center)[myvar, below=0cm of generator]{};
\node(sample)[myvar, right=1cm of generator]{virtual $\tilde{\vx}$};
\node(instance)[myvar, below=0.5cm of sample]{\phantom{a}sample $\vx$};
\node(noise)[myvar, left=1cm of generator]{$\vz + \Delta_\vz$};
\node(latent)[myvar, below=0.5cm of noise]{latent $\vz$};
\node(circ)[left=8cm of center,inner sep=0pt, outer sep=0pt]{$\bullet$};
\draw[dashed] (circ) circle (1cm); 
\draw[dashed,thick,bend right=60,myarw] (circ) -- ($(circ)+(50:1cm)$) node(pert)[auto]{$\bullet$};
\node(eps) at ($(circ) + (30:0.5cm)$){$\epsilon$};
\node(l2)[myvar, right=-0.1cm of pert] {$\vz + \mathcal{P}(\vz,\epsilon)$};
\node(l1)[myvar, below=-0.1cm of circ] { $\vz$};
\draw[myarw, bend left=70] (latent.west) to node[auto] {perturbation $\Delta_\vz$}(noise.west); 
\draw[myarw, dashed, bend right=70] (instance.east) to node[auto, right, text width=3.2cm] {on-manifold/semantic data augmentation}(sample.east); 
\draw[myarw] (noise.east) to (generator.west);
\draw[myarw] (generator.east) to (sample.west);
\draw[myarw] (instance.west) to (inverter.east);
\draw[myarw] (inverter.west) to (latent.east);
\node[auto, text width=9cm, below right=0.8cm and 0.5cm of circ, teal] {\textbf{Adversarial Perturbations} are adaptively computed by using the feedback $\nabla \mathcal{L}(\mathcal{F}^{-1}(z))$ from a classifier loss $\mathcal{L}$};
\node[auto, below right=1cm and 1.5cm of circ](ours) {};
\draw[myarw, bend right=30, teal] (ours) to (eps);

\end{tikzpicture}  
\caption{\textbf{Data augmentation via perturbation in the latent space}. Given a data sample $\vx$, natural on-manifold data augmentations are generated by perturbing the encoded $\vz = \mathcal{F}(\vx)$ in latent space, and decoding the perturbed $\vz+\Delta_\vz$. \textbf{Adversarial perturbations} require access to the loss function $\mathcal{L}$ to either find samples that are misclassified, or most difficult for the current model parameters.
}
\label{fig:schema1}
\end{figure*}

Most popular generative models for computer vision tasks are Variational Autoencoders~\cite[VAEs, ][]{kingma2014vae} or Generative Adversarial Networks~\cite[GANs, ][]{goodfellow2014generative}.

GANs are widely used in deep learning mainly due to their impressive sample quality, as well as efficient sampling. Nonetheless, by construction, these methods do not provide an invertible mapping from an image $\vx$ to its latent representation $\vz$, nor estimating its likelihood under the implicitly learned data distribution $p_{\mathcal{X}}(\vx)$, except with significant additional compromises \cite{kilcher2017generator}. Moreover, despite the notable progress, designing a stable two-player optimization method remains an active research area~\cite{chavdarova2021taming}.

VAEs, on the other hand, seemingly resolve these two problems as this class of algorithms is both approximately invertible and notably easier to train. However, VAEs are trained via maximizing a bound on the marginal likelihood and provide only approximate evaluation of $p_{\mathcal{X}}(\vx)$.
Moreover, due to their worse sample quality relative to GANs, researchers propose combining the two~\cite{larsen2016,rosca2017variational}---making their performance highly sensitive to their hyperparameter tuning.

In contrast, normalizing flows:
\begin{enumerate*}[series = tobecont, itemjoin = \quad, label=(\roman*)]
\item perform \emph{exact} encoding and decoding---due to their construction (see above, and also the illustration in Figure~\ref{fig:schema_diff}),  
\item are highly expressive,
\item are efficient to sample from,  as well as to evaluate $p_{\mathcal{X}}(\vx)$,
\item are straightforward to train, and 
\item they have useful latent representation---due to their immediate mapping from image  to latent representation.
\end{enumerate*}

In summary, apart from the obvious benefit of fast encoding and decoding when performing latent-space perturbations, to guarantee that small latent-space perturbations will not modify the sample's label, the most prominent characteristic of normalizing flows is their \textit{exact latent variable inference}. After presenting our method for perturbations in latent space and our experimental results, we further discuss the advantages of normalizing flows in~\S\ref{sec:discussion}.

\section{Perturbations in Latent Space}\label{sec:latent_attacks}

\begin{figure*}
\centering
\begin{tabular}{ccccc}
Test sample &  Randomized-LA  & Difference  & Adversarial-LA  & Difference \\
 $\vx$ & $\tilde\vx_{rand}$ &  $\tilde\vx_{rand} -\vx$  & $\tilde\vx_{adv}$ &  $\tilde\vx_{adv} -\vx$ \\[0.3ex] 
\includegraphics[height=5em,width=5em]{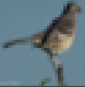} &
\includegraphics[height=5em,width=5em]{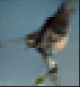}  & 
\includegraphics[height=5em,width=5em]{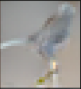}  & 
\includegraphics[height=5em,width=5em]{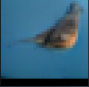} &
\includegraphics[height=5em,width=5em]{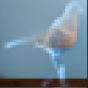}\\[0.3ex] 

\includegraphics[height=5em,width=5em]{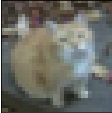} &
\includegraphics[height=5em,width=5em]{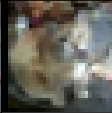}  & 
\includegraphics[height=5em,width=5em]{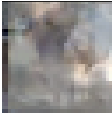}  & 
\includegraphics[height=5em,width=5em]{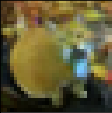} &
\includegraphics[height=5em,width=5em]{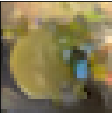}\\[0.3ex] 

\includegraphics[height=5em,width=5em]{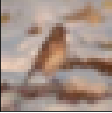} &
\includegraphics[height=5em,width=5em]{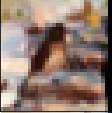}  & 
\includegraphics[height=5em,width=5em]{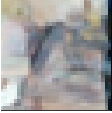}  & 
\includegraphics[height=5em,width=5em]{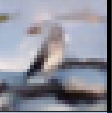} &
\includegraphics[height=5em,width=5em]{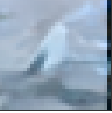}\\[0.3ex] 

\includegraphics[height=5em,width=5em]{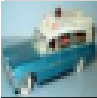} &
\includegraphics[height=5em,width=5em]{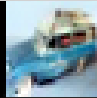}  & 
\includegraphics[height=5em,width=5em]{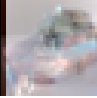}  & 
\includegraphics[height=5em,width=5em]{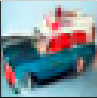} &
\includegraphics[height=5em,width=5em]{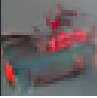}\\[0.3ex] 
\end{tabular}
\caption{\textbf{Illustrative results of our latent-space perturbations.}
The models are trained on \textbf{CIFAR-10}. 
The first column depicts randomly selected samples from the test set.
We depict the outputs obtained with Eq.~\ref{eq:r_la} and Eq.~\ref{eq:a_la} as well as their differences with the test samples.
By observing the differences, we see that the added perturbations depend on the semantic content of the input image. See~\S\ref{subsec:images} for further discussion.
}
\label{fig:samples_cifar10}
\end{figure*}

The invertibility of normalizing flows enables bidirectional transitions between image and latent spaces, see above~\S\ref{sec:nf}. This, in turn, allows for applying perturbations directly in the latent space rather than image space. We recall that we denote by 
 $\mathcal{F}: \mathcal{X} \rightarrow \mathcal{Z}$ 
a trained normalizing flow, mapping from data manifold $\mathcal{X}$ to latent space $\mathcal{Z}$.
Given a perturbation function $\mathcal{P}: \mathcal{Z} \rightarrow \mathcal{Z}$, defined over the latent space, we define its counterpart in image space as $\mathcal{F}^{-1}(\mathcal{P}(\mathcal{F}(\vx)))$.

Our goal is to define latent perturbation function $\mathcal{P}(\cdot)$ such that we obtain identity-preserving semantic modifications over the original image $\vx$ in the image domain.
To this end, we limit the structure of possible $\mathcal{P}$ in two ways.
Firstly, we directly consider incremental perturbations of the form $\vz + \mathcal{P}(\vz)$.
Secondly, we use an extra $\epsilon$ parameter to control the size of perturbation allowed (see illustration in Figure~\ref{fig:schema1}).
More precisely, we have:
$$
\mathcal{F}^{-1}\big(\mathcal{F}(\vx) + \mathcal{P}(\mathcal{F}(\vx), \epsilon)\big) \,.
$$
For brevity, we refer to $\mathcal{P}$ as \emph{latent attacks} (LA), and we consider two variants described below.

\subsection{Randomized Latent Attacks}

At training time, given a datapoint $\vx_i$, with $1 \leq i \leq N$, using the trained normalizing flow we obtain its corresponding latent code $\vz_i = \mathcal{F}(\vx_i)$. 

Primarily, as perturbation function we consider a simplistic Gaussian noise in the latent space:
\begin{align} \tag{R--LA} \label{eq:r_la}
    \mathcal{P}_{rand}(\cdot, \epsilon) = \epsilon \cdot \mathcal{N}(0,\mathbf{I}) \,,
\end{align}
which is independent from $\vz_i$.
Any such distribution around the original $\vz_i$ is equivalent to sampling from the learned manifold.
In this case, the normalizing flow \emph{pushes forward} this simple Gaussian distribution centered around $\vz_i$ to a distribution on the image space around $\vx_i = \mathcal{F}^{-1}(\vz_i)$.
Thus, sampling from the simple prior distribution $\mathcal{N}(0,\mathbf{I})$ is equivalent to sampling from a complex conditional distribution around the original image over the data manifold.

We also define norm truncated versions as follows:
$$\mathcal{P}_{rand}^{\ell_p}(\cdot, \epsilon) = \Pi(\epsilon \cdot \mathcal{N}(0,\mathbf{I})) \,,
$$
where $\ell_p$ denotes the selected norm, \textit{e.g.}, $\ell_2$ or $\ell_\infty$. For $\ell_2$ norm, $\Pi$ is defined as $\ell_2$ norm scaling, and for $\ell_\infty$, $\Pi$ is the component-wise clipping operation defined below:
\begin{equation*}
(\Pi(\vx))_i := \max(-\epsilon, \min(+\epsilon, \vx_i)) \,.
\end{equation*}

\subsection{Adversarial Latent Attacks}
Analogous to the above randomized latent attacks, at train time, given a datapoint $\vx_i$ and it's associated label $l_i$, with $1 \leq i \leq N$, using the trained normalizing flow we obtain its corresponding latent code $\vz_i = \mathcal{F}(\vx_i)$. 

We search for $\Delta_{\vz_i} \in \mathcal{Z}$ such that the loss obtained of the generated image $\tilde\vx_i =  \mathcal{F}^{-1} (\vz_i + \Delta_{\vz_i}) $ is maximal:
\begin{align}  \notag
\Delta_{\vz_i}^\star &= \argmax_{\|\Delta_{\vz_i}\|_{l_p} \leq \epsilon} \mathcal{L}_{\theta}(\mathcal{F}^{-1}(\vz_i + \Delta_{\vz_i}), l_i ) \,,  \\
\mathcal{P}_{adv}^{\ell_p}(\vz_i, \epsilon) &= \Delta_{\vz_i}^\star  \,, \label{eq:a_la} \tag{A--LA} 
\end{align}
where $\mathcal{L}_{\theta}$ is the loss function of the classifier, and $\ell_p$ denotes the selected norm, \textit{e.g.},\ $\ell_2$ or $\ell_\infty$.

In practice, we define the number of steps $k$ to optimize for $\Delta_{\vz_i}^\star \in \mathcal{Z}$, as well as the step size $\alpha$~\cite[similar to ][]{stutz2019disentangling, wong2020learning}, and we have the following procedure:

\begin{itemize}
    \item Initialize a random $\Delta_{\vz_i}^0$ with $\|\Delta_{\vz_i}^0\|_{\ell_p} \leq \epsilon$.
    \item Iteratively update $\Delta_{\vz_i}^j$ for $j=1, \dots,k$ number of steps with step size $\alpha$ as follows:
$$\Delta_{\vz_i}^j = \Pi\Big(\Delta_{\vz_i}^{j-1} + \alpha \cdot \frac{\nabla\mathcal{L}_{\theta}(\mathcal{F}^{-1}(\vz_i + \Delta_{\vz_i}^{j-1}), l_i)}{\|\nabla\mathcal{L}_{\theta}(\mathcal{F}^{-1}(\vz_i + \Delta_{\vz_i}^{j-1}), l_i)\|_{\ell_p}}\Big)$$
where $\Pi$ is the projection operator that ensures condition $\|\Delta_{\vz_i}^j\|_{\ell_p} \leq \epsilon$ and gradient is with respect to $\Delta_{\vz_i}^{j-1}$.
    \item Output $\mathcal{P}_{adv}(\vz_i, \epsilon) = \Delta_{\vz_i}^k$
\end{itemize}

For the case of $\ell_\infty$, we replace normalization of gradient with $sign(\cdot)$ operator, \textit{i.e.}:
$$\Delta_{\vz_i}^j = \Pi\Big(\Delta_{\vz_i}^{j-1} + \alpha \cdot sign\big(\nabla\mathcal{L}_{\theta}(\mathcal{F}^{-1}(\vz_i + \Delta_{\vz_i}^{j-1}), l_i)\big)\Big)$$
and use component-wise clipping for projection, which is equivalent to the standard $\ell_\infty$-PGD adversarial attack of~\citet{madry2017towards}.

Similarly, as the normalizing flow directly models the underlying data manifold, this perturbation is equivalent to a search over the \emph{on-manifold} adversarial samples \cite{stutz2019disentangling}.

\section{Experiments}\label{sec:experiments}

\paragraph{Datasets.}
We evaluate our proposed semantic perturbations on the \textbf{FashionMNIST}, \textbf{SVHN}, \textbf{CIFAR-10}, and \textbf{CIFAR-100} datasets.
See Appendix~\ref{app:results} for additional results on \textbf{MNIST}. For experiments on restricted datasets, \textit{e.g.}, 5\% of CIFAR-10, we always use the same sample set for a fair comparison.

\paragraph{Models.} For FashionMNIST, we use a conditional $12$-step normalizing flow based on \textit{Glow} coupling blocks and a convolutional network of approximately $100K$ parameters, as in \cite{stutz2019disentangling}. For experiments on SVHN and CIFAR-10/100, we use Glow~\cite{kingma2018glow} and ResNet-18~\cite{resnet}, respectively. See Appendix~\ref{app:iml_details} for further details on the implementation.

\paragraph{Metrics.}
\vspace{-0.05cm}
To evaluate the classifier's generalization, we use standard test accuracy. Adopting from the literature on GANs, we use Fr\'echet Inception Distance~\cite[FID, lower is better, see Appendix~\ref{app:metrics},][]{heusel2017gans} to measure the similarity between the CIFAR-10 training data and samples produced by our latent perturbations.

\paragraph{Methods.}
\vspace{-0.05cm}
We compare the following methods:
\begin{enumerate*}[series = tobecont, itemjoin = \quad, label=(\roman*)]
\item \textbf{standard}--classical training with no attacks,
\item \textbf{Image-space PGD}: Projected Gradient Descent as an image-space, adversarial perturbation baseline \cite{madry2017towards},
\item \textbf{VAE-GAN}~\cite{stutz2019disentangling}--on-manifold perturbation method that uses VAE-GANs,
\item \textbf{Cutout}~\cite{devries2017cutout}--input masking,
\item \textbf{Mixup}~\cite{zhang2018mixup}--data-agnostic data augmentation routine,
\item \textbf{Randomized-LA (ours)}--randomized latent attacks using normalizing flow, as well as
\item \textbf{Adversarial-LA (ours)}--adversarial latent attacks using normalizing flow,
\end{enumerate*}
where \textit{ours} are described in~\S\ref{sec:latent_attacks} and the rest of the methods in~\S\ref{sec:related_works}. For brevity, PGD, Randomized-LA and Adversarial-LA are sometimes denoted with $\mathcal{P}_{pgd}$, $\mathcal{P}_{rand}$ and $\mathcal{P}_{adv}$, respectively. 

\subsection{Generalization on CIFAR-10}
\label{subsec:cifar10}

\begin{table}[t]\centering
\begin{tabular}{lcc}
\toprule  
\textbf{Method}                             & \textbf{Low-data}     & \textbf{Full-set}  \\ \midrule
Standard (no DA)                            & $49.8$ & $89.7$ \\
Standard $+$ common DA                    & $64.1$                & $95.2$ \\
VAE-GAN~\cite{stutz2019disentangling}   & $58.9$                      & $94.2$  \\
Cutout~\cite{devries2017cutout}             & $ 66.8$              &  $96.0$ \\
Mixup~\cite{zhang2018mixup}                 & $73.4$               & $95.9$ \\
Randomized-LA                        & $70.1$               & $96.3$ \\
Adversarial-LA                       & $\mathbf{80.4}$               & $\mathbf{96.6}$ \\
\bottomrule
\end{tabular}
\vspace{0.2cm}
\caption{Test accuracy ($\%$) on \textbf{CIFAR-10}, in the \emph{low-data regime} compared to the \emph{full train set}.
For the former, we use $5\%$ and $100\%$ of the training and test set, respectively.
In addition to standard training, we consider standard training with commonly used data augmentations (DA) in the image space, which includes rotation and horizontal flips~\cite{zagoruyko_wide_2016}, as well as more recent \textit{Cutout} \cite{devries2017cutout} and \textit{Mixup}~\cite{zhang2018mixup} methods.
See~\S\ref{subsec:cifar10} for a discussion.
}
\vspace{-0.5em}
\label{tab:lowcifar10}
\end{table}

We are primarily interested in the performance of our perturbations in the low-data regime when using only a small subset of CIFAR-10 as the training set. We train ResNet-18 classifiers on only $5\%$ of the full training set and evaluate models on the full test set. 
We compare our methods with some of the most commonly used data augmentations methods such as \textit{Cutout} \cite{devries2017cutout} and \textit{Mixup}~\cite{zhang2018mixup}, as well as with the VAE-GAN based approach \cite{stutz2019disentangling}.
For \cite{stutz2019disentangling}, we use the authors' implementation, and their default parameters for CelebA dataset, see Appendix~\ref{app:iml_details} for details.
For \cite{devries2017cutout}, we report the best test accuracy observed among a grid search on the learning rate $ \eta \in \{0.1,0.01\}$. Similarly, for \cite{zhang2018mixup}, we report the best accuracy among grid search on learning rate $\eta \in \{0.1,0.01\}$ and mixup coefficient $\lambda \in \{.1,.2,.3,.4,1.0\}$.
For Randomized-LA, we use $\ell=\ell_\infty, \epsilon=0.25$, and for Adversarial-LA, we use $\ell=\ell_2, \epsilon=1.0, \alpha=0.5, k=3$.

\begin{figure}[h]
    \centering
    \vspace{-0.32cm}
    \includegraphics[width=\columnwidth]{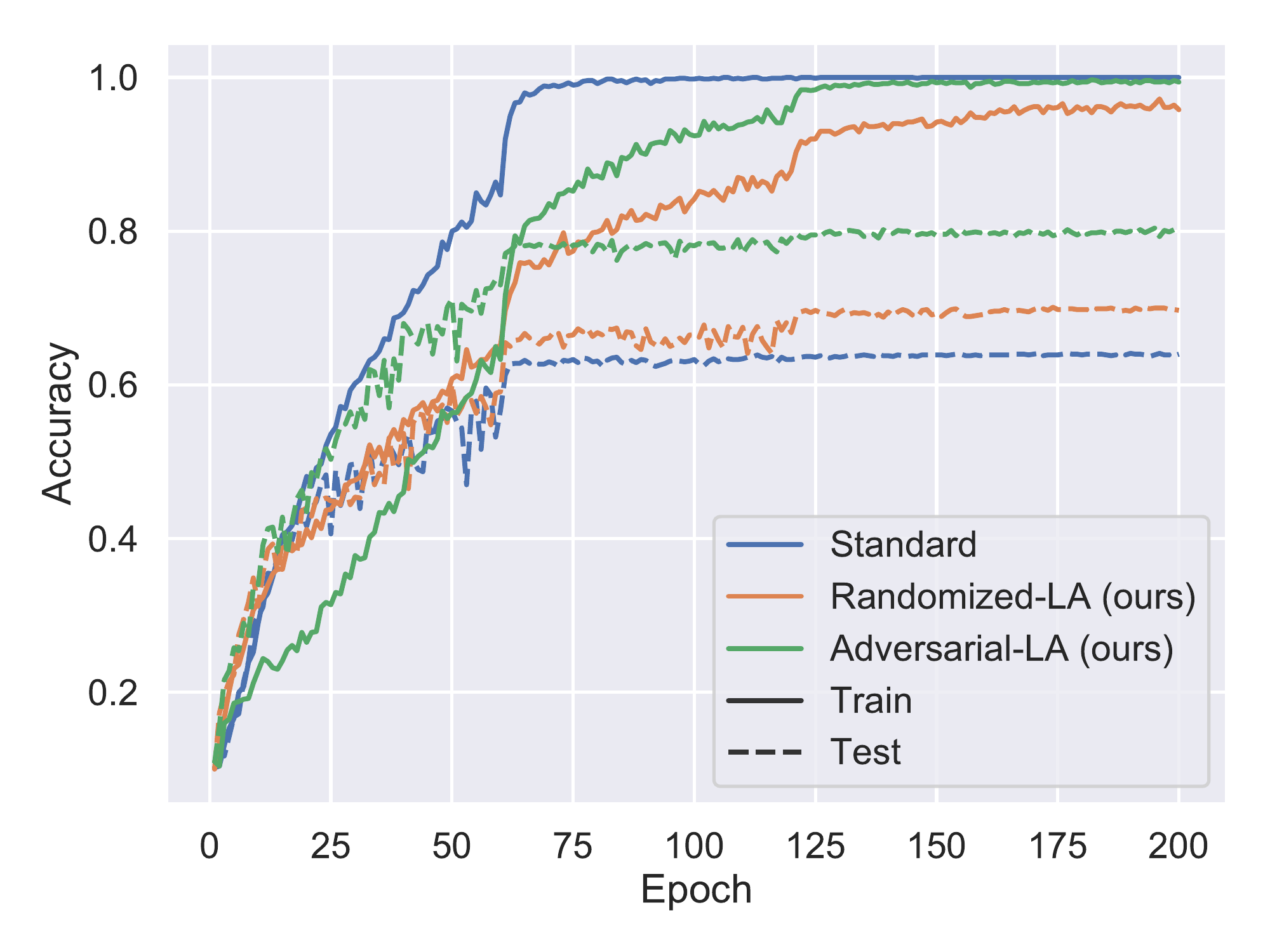}
    \vspace{-1.5em}
    \caption{Accuracy when training on $5\%$ of the \textbf{CIFAR-10} dataset, and testing on its full test set.
    See~\S\ref{subsec:cifar10} for a discussion.}
    \label{fig:cifar10_curves}
\end{figure}

Table~\ref{tab:lowcifar10} summarizes our generalization experiments in the low data regime---using only $5\%$ of CIFAR-10 for training, compared to the full CIFAR-10 training set. Figure~\ref{fig:cifar10_curves} depicts the train and test accuracy throughout the training. Both Randomized-LA and Adversarial-LA notably outperform the standard training baseline. In particular, we observe that
\begin{enumerate*}[series = tobecont, itemjoin = \quad, label=(\roman*)]
\item our simplistic Randomized-LA method already outperforms some recent strong data augmentation methods, and  
\item Adversarial-LA achieves \emph{best} test accuracy for both low-data and full-set regimes.
\end{enumerate*}
See~\S\ref{subsec:stutz} below for additional benchmarks with VAE-GAN~\cite{stutz2019disentangling}.

\subsection{Transfer Learning Experiments}\label{subsec:cifar100}

To further analyze potential applications of our normalizing flow based latent attacks to real-world use cases, we study if a normalizing flow pre-trained on a large dataset can be used for training classifiers on a different, smaller dataset.
In particular, we use CIFAR-10 to train the normalizing flows and then our latent attacks to train a classifier on $10\%$ and $5\%$ of the CIFAR-100 and SVHN training datasets, respectively.

Table~\ref{tab:lowcifar100} shows our results for CIFAR-100 using a selection of latent attacks. Randomized-LA and Adversarial-LA achieve $16\%$ and $24\%$ improvements over the standard baseline. The results indicate that normalizing flows are capable of transferring useful augmentations learned from CIFAR-10 to CIFAR-100.

Table~\ref{tab:lowsvhn} shows our results for SVHN. To provide a baseline on the effect of using different datasets for normalizing flows and classifiers, we also provide results with pre-training on SVHN. Latent attacks transferred from CIFAR-10 achieve superior performance to direct pre-training on SVHN, indicating that transferring augmentations across datasets is indeed a promising direction.

\begin{table}
\centering
\begin{tabular}{lc}
\toprule  
\textbf{Perturbation}                              & \textbf{Accuracy}  \\ \midrule
Standard                                     & $36.4$ \\
Randomized-LA, $\ell{=}\ell_\infty,\epsilon{=}.2$               & $39.7$ \\
Randomized-LA, $\ell{=}\ell_\infty,\epsilon=.3$               & $41.0$ \\
Randomized-LA, $\ell{=}\ell_2,\epsilon=10$                             & $40.4$ \\
Randomized-LA, $\ell{=}\ell_2,\epsilon=20$                             & $\mathbf{42.3}$ \\
Adversarial-LA, $\ell{=}\ell_2,\alpha{=}.5,k{=}3$            & $\mathbf{45.0}$ \\
\bottomrule  
\end{tabular}
\vspace{0.2cm}
\caption{Test accuracy ($\%$) on \textbf{CIFAR-100}, in the \emph{low-data regime}, where we use $10\%$ of the training set and the full test set. The normalizing flow used is trained on \textbf{CIFAR-10}.
}
\vspace{-1.0em}
\label{tab:lowcifar100}
\end{table}

\begin{table}[h]
\centering
\begin{tabular}{llc}
\toprule
\textbf{NF} & \textbf{Perturbation} & \textbf{Accuracy} \\
\midrule
-- & Standard                    & $81.2$\\
\midrule
\multirow{2}{*}{{SVHN}} & $\mathcal{P}_{rand}^{\ell_2}$, \small{$\epsilon{=}15.$}             & $84.9$\\
& $\mathcal{P}_{adv}^{\ell_2}$, \small{$\epsilon{=}.5, \alpha{=}.25, k{=}2$}             & $86.9$ \\
\midrule
\multirow{2}{*}{{CIFAR-10}} & $\mathcal{P}_{rand}^{\ell_2}$, \small{$\epsilon{=}15.$} & $\mathbf{90.0}$ \\
& $\mathcal{P}_{adv}^{\ell_2}$, \small{$\epsilon{=}.3, \alpha{=}.15, k{=}2$} & $\mathbf{90.5}$\\
\bottomrule  
\end{tabular}
\vspace{0.2cm}
\caption{Test accuracy ($\%$) on \textbf{SVHN}, in the \emph{low-data regime}, where we use $5\%$ of the training set and the full test set. Comparison of normalizing flows trained on \textbf{CIFAR-10}, versus \textbf{SVHN}.
}
\label{tab:lowsvhn}
\end{table}

\subsection{Additional Comparison with VAE-GAN}\label{subsec:stutz}

\begin{figure}
    \vspace{-0.5em}
    \centering
    \includegraphics[width=\columnwidth]{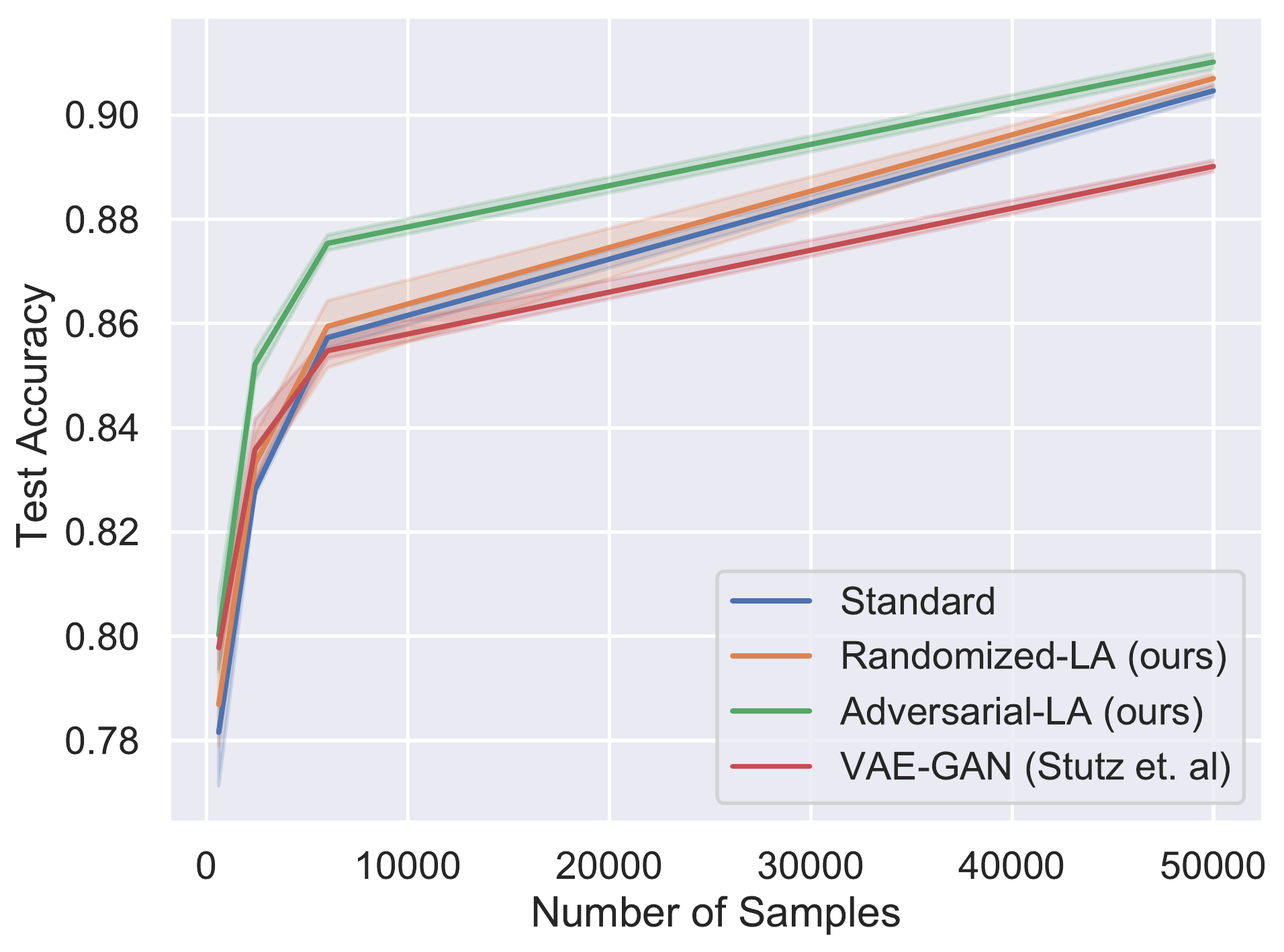}
    \vspace{-1.0em}
    \caption{Test accuracy ($\%$) on full test set, for a varying number of \textit{training} samples, on the \textbf{FashionMNIST} dataset.
    To replicate the setup of VAE-GAN~\cite{stutz2019disentangling}, only a portion of the dataset (x-axis) is used to train the classifier, while the corresponding generative model is trained on the full dataset.
    We run each experiment with three different random seeds and report the mean and standard deviation of the test accuracy. See~\S\ref{subsec:stutz}.
    }
    \label{fig:fmnist_curves}
    \vspace{-0.5em}
\end{figure}

Following~\citet{stutz2019disentangling}, we study the performance of our latent perturbation-based training strategies in varying settings, starting from low-data regime to full-set.
For the VAE-GAN results, we use the source code provided by the authors, while using their default hyperparameters for the same dataset. For our methods, we reproduce the same classifier and hyperparameter setup.
For Randomized-LA, we use $\ell=\ell_\infty, \epsilon=0.15$, and for Adversarial-LA, we use $\ell=\ell_\infty, \epsilon=0.05, \alpha=0.01, k=10$.

Figure~\ref{fig:fmnist_curves} shows our average results for $3$ runs with training sizes in $\{600,2400,6000,50000\}$. We observe that Randomized-LA performs comparatively to the standard training baseline, whereas Adversarial-LA outperforms the standard baseline across all train set sizes. Note that the difference to the standard baselines shrinks as we increase the number of samples available to the classifiers.  

In line with our results, \citet{stutz2019disentangling} report diminishing performance gains for increasingly challenging datasets such as FashionMNIST to CelebA, when using therein VAE-GAN based approach. One potential cause could be the \emph{approximate} encoding and decoding mappings or sensitivity to hyperparameter tuning.
Relative to VAE-GAN, normalizing flows have significantly fewer hyperparameters, see Appendix~\ref{app:hyperparams}. 
Indeed, our results support the numerous appealing advantages of normalizing flows for latent-space perturbations and indicate that they have a better capacity to produce useful augmented training samples.

\subsection{Analysis of Generated Images}
\label{subsec:images}

\begin{table}
    \centering
    \begin{tabular}{ll}
    \toprule
    \textbf{Model or Perturbation}  &  \textbf{FID} \\
    \midrule
    \textit{Baseline: GANs} &\\ 
    DCGAN \cite{heusel2017gans}           &  36.9 \\
    WGAN-GP \cite{heusel2017gans}         &  24.8  \\
    BigGAN \cite{brock2018large}          &  14.73  \\
    StyleGAN \cite{karras2020training}    &  2.92  \\
    \hdashline \vspace{-0.3cm} \\
    \textit{Baseline: image-space} &\\ 
    PGD \cite{madry2017towards}, $\ell{=}\ell_\infty,\epsilon{=}.03,\alpha{=}.008,k{=}10$ & 23.61 \\     \midrule
    \textit{Ours: latent-space} &\\
    Randomized-LA, $\ell{=}\ell_\infty,\epsilon{=}.25$ & 3.71 \\     Adversarial-LA, $\ell{=}\ell_2,\epsilon{=}1.,\alpha{=}.5,k{=}3$ & 3.65 \\     \bottomrule
    \end{tabular}
    \vspace{0.2cm}
    \caption{FID scores (lower is better) of generated samples of GANs, 
    image-space PGD perturbations, and our Randomized-LA and Adversarial-LA methods.
    For PGD and Adversarial-LA perturbations which use a classifier, we use the same standardly trained ResNet-18. See~\S\ref{subsec:images}.}
    \label{tab:fid}
\end{table}

Figure~\ref{fig:samples_cifar10} depicts samples of our Randomized-LA and Adversarial-LA methods. Primarily, in contrast to random image-space perturbations, we observe that both Randomized-LA and Adversarial-LA yield perturbations dependent on the semantic content of the input image.
Interestingly, one could argue that Adversarial-LA further masks potential \textit{shortcuts}  that the classifier may learn, for \textit{e.g.}, by masking the windows, it forces the classifier to, in fact, learn the shape of a car.
Moreover, we observe that relative to image-space perturbations, latent attacks produce samples that are semantically closer to the CIFAR-10 training set---see Table~\ref{tab:fid} for FID scores, and at the same time, more distinct in image space---see Table~\ref{tab:distance}.

\begin{table}
    \centering
   \begin{tabular}{lcc}
   \toprule
   \textbf{Perturbation}  &   $\ell_2$ in $\mathcal{X}$   &  $\ell_\infty$ in $\mathcal{X}$ \\
   \midrule
   \textit{Baseline: image-space} &&\\
    $\mathcal{P}_{pgd}^{\ell_\infty}$,  \small{$\epsilon=.03,\alpha=.008, k=10$} & $1.13$ & $0.03$ \\
    $\mathcal{P}_{pgd}^{\ell_2}$, \small{$\epsilon=2.,\alpha=.5,k=10$}        & $1.98$ & $0.15$ \\
   \midrule
   \textit{Ours: latent-space} &&\\
   $\mathcal{P}_{rand}^{\ell_\infty}$, \small{$\epsilon=.25$}                    & 4.18 & 0.41 \\
   $\mathcal{P}_{adv}^{\ell_2}$, \small{$\epsilon=1.,\alpha=.5,k=3$}          & $4.61$ & $0.44$ \\
   \bottomrule
   \end{tabular}
   \vspace{0.2cm}
   \caption{Average $\ell_2$ and $\ell_\infty$ size of perturbations computed in image space $\mathcal{X}$ using CIFAR-10 test samples. For PGD and Adversarial-LA perturbations which use a classifier, we use the same standardly trained ResNet-18.}
   \label{tab:distance}
   \vspace{-0.3cm}
\end{table}

\subsection{Robustness against Latent Attacks}
\label{subsec:robustness}

\begin{table}
    \centering
   \begin{tabular}{llccc}
   \toprule
   \textbf{Attack} 
   & \textbf{Trained Perturbation}    &  \textbf{Acc.} & \textbf{Drop} \\
   \midrule
   \multirow{4.5}{*}{$\mathcal{P}_{rand}^{\ell_\infty}$} & Standard & 90.5 & 4.8 \\    & $\mathcal{P}_{pgd}^{\ell_\infty}$, \small{$\epsilon=.03,\alpha=.008, k=10$} & 76.9 & 9.4 \\    \cmidrule{2-4}
   & $\mathcal{P}_{rand}^{\ell_\infty}$, \small{$\epsilon=.25$} & 94.1   & 2.1 \\    & $\mathcal{P}_{adv}^{\ell_2}$, \small{$\epsilon=1., \alpha=.5, k=3$} & 94.6 & 2.0 \\    \midrule
   \multirow{4.5}{*}{$\mathcal{P}_{adv}^{\ell_2}$} & Standard & 58.8 & 38.2  \\     & $\mathcal{P}_{pgd}^{\ell_\infty}$, \small{$\epsilon=.03,\alpha=.008, k=10$} & 36.2  & 57.3 \\    \cmidrule{2-4}
   & $\mathcal{P}_{rand}^{\ell_\infty}$, \small{$\epsilon=.25$} & 71.2 & 25.9 \\    & $\mathcal{P}_{adv}^{\ell_2}$, \small{$\epsilon=1., \alpha=.5, k=3$} & 76.4 & 20.8 \\    \bottomrule
   \end{tabular}
   \caption{Robustness against our perturbations ${\mathcal{P}_{rand}^{\ell_\infty},\epsilon=.25}$ (\textbf{top}) and ${P_{adv}^{\ell_2},\epsilon=1., \alpha=.5, k=3}$ (\textbf{bottom}) on the \textbf{CIFAR-10} dataset.
   \textbf{Trained Perturbation}: the \emph{training-time} perturbation used to train the model; \textbf{Drop} the drop in test accuracy with latent perturbations \emph{relative} to the accuracy on CIFAR-10 test samples.
   }
    \vspace{-0.7em}
    \label{tab:adv}
\end{table}

In Table~\ref{tab:adv}, we evaluate the robustness of classifiers against our latent attacks and observe that both standard and image-space adversarial training suffer from a significant loss of performance against Adversarial-LA. Combined with observations from~\S\ref{subsec:images}, this indicates that our adversarial latent attack is a novel approach to generate \emph{realistic} adversarial samples. Interestingly, classifiers trained with image-space adversarial perturbations are more prone to large accuracy drops than standardly trained classifiers.
Additionally, although the classifiers trained with our perturbations are robust to Randomized-LA, they are not fully robust to Adversarial-LA, suggesting the possibility of further improving generalization using latent attacks.

\section{Discussion}\label{sec:discussion}

\paragraph{Exact Coding.}
As formalized in~\S\ref{sec:nf}, normalizing flows can perform exact encoding and decoding by their construction. That is, the decoding operation is exactly the reverse of the encoding operation. Any continuous encoder maps a neighborhood of a sample to some neighborhood of its latent representation. However, the invertibility of normalizing flows also maps \emph{any neighborhood of latent code to a neighborhood of the original sample}. 
In principle, this property also holds for off-manifold samples and may explain the effectiveness of our methods in transferring augmentations.

\paragraph{Increasing Dataset Size.}
The primary advantage of exact coding is that the generated samples via latent perturbations improve the generalization of classifiers, as shown in~\S\ref{subsec:cifar10}. 
To understand why this occurs, consider the limit case $\epsilon\to 0$ for a latent perturbation. Assuming a numerically stable normalizing flow, we recover the original data samples, hence the training distribution. As we increase $\epsilon$, this distribution grows around each data point. Thus, by increasing $\epsilon$, we add further plausible data points to our training set, as long the learned latent representation is a good approximation of the underlying data manifold. This does not necessarily hold for approximate methods due to inherent \textit{decoder noise}.

\paragraph{Controllability.}
In~\S\ref{sec:latent_attacks}, we introduced two variants of latent perturbations that define different procedures around the latent code of the original sample. Each variant employs a normalizing flow to efficiently map a complex on-manifold objective to a local objective in the latent space. The randomized latent attack defines a sampling operation on the data manifold, and the adversarial latent attack, a stochastic search procedure to find on-manifold samples attaining high classifier losses.
In principle, any other on-manifold objectives may also utilize such mappings to the latent space and, potentially, use the density provided by the normalizing flow to enforce strict checks for on-manifold data points. Moreover, conditional normalizing flows may achieve more expressive, class-specific augmentations and control mechanisms.

\paragraph{Compatibility with Data Augmentations.}
It is important to note that our method is orthogonal to image-space data augmentation methods. In other words, we can train normalizing flows with commonly used data augmentations.
As observed in Figure~\ref{fig:samples_cifar10}, trained models can apply some of the training-time augmentations to CIFAR-10 test samples. This allows us to encode and decode \emph{augmented} samples as well as original samples of CIFAR-10.
Additionally, we can use \citet{devries2017cutout, zhang2018mixup} concurrently with our latent perturbations to train classifiers. 

\section{Conclusion}\label{sec:conclusion}

Motivated by the numerous advantages of normalizing flows, we propose flow-based latent perturbation methods to augment the training datasets to train classifiers.  
Our extensive empirical results on several real-world datasets demonstrate the efficacy of these perturbations for improving generalization both in full and low-data regimes.
In particular, these perturbations can increase sample efficiency in low-data regimes and, in practice, reduce labeling efforts.

Further directions include 
\begin{enumerate*}[series = tobecont, itemjoin = \quad, label=(\roman*)]
\item decoupling the effects of exact coding from any modeling gains through ablation studies, as well as 
\item combining image and latent-space augmentations.
\end{enumerate*}

\subsubsection*{Acknowledgments}
TC was funded in part by the grant P2ELP2\_199740 from 
the Swiss National Science Foundation.
The authors would like to 
thank Maksym Andriushchenko and Suzan Üsküdarlı for insightful feedback and discussions.

{\small
\bibliographystyle{ieee_fullname}
\bibliography{main}
}

\appendix
\newpage
\onecolumn
\section{Details on the implementation}\label{app:iml_details}

In this section, we list all the details of the implementation.

\vspace{-.25cm}
\paragraph{Source Code.} Our source code is provided in this repository: \url{https://github.com/okyksl/flow-lp}.

\subsection{Architectures}\label{app:arch}

\paragraph{Generative model (NF) architecture.}

We use \textit{Glow}~\cite{kingma2018glow} for the normalizing flow architecture. For the MNIST~\cite{lecun_gradient-based_1998} and FashionMNIST~\cite{xiao2017/online} experiments, we use a conditional, 12-step, Glow-coupling-based architecture similar to~\cite{ardizzone2019guided}. See Table~\ref{tab:mnist_flow} for the details. For the CIFAR-10/100~\cite{krizhevsky_learning_2009} and SVHN~\cite{netzer2011reading} experiments, we use the original Glow architecture described in~\cite{kingma2018glow}, \emph{i.e.}, 3 scales of 32 steps each containing activation normalization, affine coupling and invertible 1$\times$1 convolution. We adapt an existing PyTorch implementation in \footnote{\url{https://github.com/chrischute/glow}} to better match the original Tensorflow implementation in \footnote{\url{https://github.com/openai/glow}}. For more details on multi-scale architecture in normalizing flows, see~\cite{dinh2016density}.

\begin{table*}[!h]\centering		 \begin{minipage}[b]{0.29\hsize}\centering  \begin{tabular}{@{}c@{}}\toprule
\textbf{Generative Model}\\\toprule
\textit{Input:} $\vx \in \R^{784}, \vy \in \R^{10}$ \\  \hdashline 
 GLOWCouplingBlock \\
 PermuteRandom \\
\hdashline
 $\cdots$ \\
 $\times10$ \\
 $\cdots$ \\
\hdashline
 GLOWCouplingBlock \\
 PermuteRandom \\
\bottomrule \vspace{.1cm}
\end{tabular}
\end{minipage} \hfill
\begin{minipage}[b]{0.4\hsize}\centering  \begin{tabular}{@{}c@{}}\toprule
\textbf{GLOWCouplingBlock}\\\toprule
\textit{Input:} $\vx \in \R^{784}, \vy \in \R^{10}$ \\ [0.3ex]
 \hdashline
 split $\vx\rightarrow\vx_1,\vx_2$ ($784\rightarrow{392},{392}$) \\ [0.3ex]
 \hdashline
 subnet $\vx_2 \oplus \vy \rightarrow \mathbf{s}_1, \mathbf{t}_1$ ($402 \rightarrow 392,392$) \\ [0.3ex]
 affine coupling $\vx_1, \mathbf{s}_1, \mathbf{t}_1 \rightarrow \vz_1$ ($3{\times}392 \rightarrow 392$)\\ [0.3ex]
 \hdashline
 subnet $\vz_1 \oplus \vy \rightarrow \mathbf{s}_2, \mathbf{t}_2$ ($402 \rightarrow 392,392$) \\ [0.3ex]
 affine coupling $\vx_2, \mathbf{s}_2, \mathbf{t}_2 \rightarrow \vx_2^{'}$ ($3{\times}392 \rightarrow 392$)\\ [0.3ex]
  \hdashline
 concat. $\vz_1 \oplus \vz_2$ ($392,392 \rightarrow 784$) \\ [0.3ex]
\bottomrule \vspace{.1cm}
\end{tabular}
\end{minipage}
\begin{minipage}[b]{0.29\hsize}\centering  \begin{tabular}{@{}c@{}}\toprule
\textbf{Subnets}\\\toprule
\textit{Input:} $\vx \in \R^{402}$ \\ [0.3ex]
 \hdashline
 linear ($402 \rightarrow 512$) \\
 ReLU \\
 linear ($512 \rightarrow 784$) \\
 split ($784 \rightarrow 392,392$) \\
\bottomrule \vspace{.1cm}
\end{tabular}
\end{minipage}
\caption{Normalizing flow architectures used for our experiments on \textbf{MNIST} and \textbf{FashionMNIST}. With $ c_{in} \rightarrow y_{out}$, we denote the number of channels of the input and output of the layer. With $\oplus$, we denote concatenation operation. We use the implementation provided in \url{https://github.com/VLL-HD/FrEIA}. For more details on affine coupling layers, see \S\ref{sec:nf}.
}
\vspace{-1.0em}\label{tab:mnist_flow}
\end{table*}

\paragraph{Classifier architecture.} For our experiments on MNIST, we use LeNet-5~\cite{lecun_gradient-based_1998} with replaced nonlinearity--instead of $\tanh$ we use $\text{ReLU}$, and we initialize the network parameters with truncated normal distribution $\sigma = 0.1$. For the FashionMNIST experiments, we use the same classifier as used in~\cite{stutz2019disentangling}. See Table~\ref{tab:mnist_classifier} for more details. For CIFAR-10/100 and SVHN, we use the ResNet-18 \cite{he2016identity} architecture as implemented in \cite{devries2017cutout, zhang2018mixup}. 
This ResNet-18 includes slight modifications over the standard ResNet-18 architecture in order to achieve better performance on CIFAR-10/100. See \footnote{\url{https://github.com/facebookresearch/mixup-cifar10}} and \footnote{\url{https://github.com/uoguelph-mlrg/Cutout}} for implementation. 
In particular, the first layer is changed to a $3\times3$ convolution with stride ${1}$ and padding ${1}$, from the original $7\times7$ convolution with stride $2$ and padding $3$. Additionally, the following max-pooling layer is removed. For CIFAR-10, we also use a similarly modified ResNet-20 \cite{he2016deep}.

\begin{table}\centering		 \begin{minipage}[b]{0.49\hsize}\centering  \begin{tabular}{@{}c@{}}\toprule
\textbf{LeNet-5}\\\toprule
\textit{Input:} $\vx \rightarrow \R^{1{\times}28{\times}28} $ \\  \hdashline 
 convolution (ker: $5{\times}5$, $1 \rightarrow 6$; stride: $1$; pad:$2$) \\
 ReLU \\
 AvgPool2d (ker: $2{\times}2$) \\
 convolution (ker: $5{\times}5$, $6 \rightarrow 16$; stride: $1$; pad:$0$) \\
 ReLU \\
 AvgPool2d (ker: $2{\times}2$) \\
 Flatten ($16{\times}5{\times}5\rightarrow400$) \\
 linear ($400\to120$)\\
 ReLU \\
 linear ($120\to84$)\\
 ReLU \\
 linear ($120\to10$)\\
 ReLU \\
\bottomrule \vspace{.1cm}
\end{tabular}
\end{minipage} \hfill
\begin{minipage}[b]{0.49\hsize}\centering  \begin{tabular}{@{}c@{}}\toprule
\textbf{CNN from \cite{stutz2019disentangling}}\\\toprule
\textit{Input:} $\vx \in \R^{1{\times}28{\times}28} $ \\  \hdashline 
 convolution (ker: $4{\times}4$, $1 \rightarrow 16$; stride: $2$; pad:$1$) \\
 Batch Normalization \\ 
 ReLU \\
 convolution (ker: $4{\times}4$, $16 \rightarrow 32$; stride: $2$; pad:$1$) \\
 Batch Normalization \\ 
 ReLU \\
 convolution (ker: $4{\times}4$, $32 \rightarrow 64$; stride: $2$; pad:$1$) \\
 Batch Normalization \\
 ReLU \\
 Flatten ($64{\times}3{\times}3\rightarrow576$) \\
 linear ($576\to100$)\\
 linear ($100\to10$)\\
\bottomrule \vspace{.1cm}
\end{tabular}
\end{minipage}
\caption{
Convolutional Neural Network (CNN) architectures used for our experiments on \textbf{MNIST} and \textbf{FashionMNIST}.
We use \textit{ker} and \textit{pad} to denote \textit{kernel} and \textit{padding} for the convolution layers, respectively. 
With $h{\times}w$, we denote the kernel size.
With $ c_{in} \rightarrow y_{out}$, we denote the number of channels of the input and output of the layer.
}\label{tab:mnist_classifier}
\end{table}

\subsection{Hyperparameters}\label{app:hyperparams}

\paragraph{Generative Models.}
For MNIST and FashionMNIST, we use the Adam~\cite{kingma2014adam} optimizer with a batch size of $100$ and learning rate of $10^{-6}$ for $100$ epochs to train normalizing flows. For CIFAR-10 and SVHN, we use the Adamax~\cite{kingma2014adam} optimizer with a learning rate of $0.0005$ and weight decay of $0.00005$. We use a warmup learning rate schedule for the first $500.000$ steps of the training. That is, the learning rate is linearly increased from $0$ to the base learning rate $0.0005$ in $500.000$ steps.

For VAE-GAN training, we run the implementation provided by authors\footnote{\url{https://github.com/davidstutz/disentangling-robustness-generalization}} with the default architectures and parameters. That is, for FashionMNIST, we use $\beta=2.75$, $\gamma=1$, $\eta=0$ and latent space size of $10$. We use the Adam optimizer with a batch size of $100$, learning rate of $0.005$, weight decay of $0.0001$ and train VAE-GANs for $60$ epochs with an exponential decay scheduling of $0.9$ for the learning rate. For CIFAR-10, we use the CelebA~\cite{liu2015faceattributes} setup provided (the only 3-channel color dataset provided) and thus use $\beta=3.0$, latent space size of $25$ and 30 epochs instead. Note that we report \emph{On-Learned-Manifold Adversarial Training} from \cite{stutz2019disentangling} which uses class-specific VAE-GANs. That is, 10 VAE-GAN architectures are trained for both {FashionMNIST} and {CIFAR-10} datasets. 

\paragraph{Discussion on Hyperparameters of Generative Models.} As normalizing flows directly optimize the log-likelihood of the data, there are no hyperparameters in their loss function. Additionally, the normalizing flows that we use have a fixed latent dimension equal to the input dimension due to their architectural design. As noted in \S\ref{subsec:stutz}, this is in contrast to VAE-GAN used in \cite{stutz2019disentangling} where the training involves optimizing separate losses for three networks (namely, encoder, decoder, and discriminator) concurrently. Coefficients called $\beta$, $\gamma$, and $\eta$ are used to scale reconstruction, decoder, and discriminator loss, respectively. Additionally, the latent size for VAE-GAN is hand-picked for each dataset.

\paragraph{Classifiers.} For MNIST, we use the Adam optimizer with a learning rate of $0.001$ and weight decay of $0.001$. We train LeNet-5 classifiers for $20$ epochs with exponential learning decay of rate $0.1$ for $10.000$ steps. For FashionMNIST, we use the training setup used in \cite{stutz2019disentangling}. That is, we use the Adam optimizer with a learning rate of $0.01$ and weight decay of $0.0001$. We train classifiers for $20$ epochs with exponential learning decay of rate $0.9$ for $500$ steps. For CIFAR-10/100, we use the training setup used in \cite{devries2017cutout, zhang2018mixup}. More precisely, we use Stochastic Gradient Descent (SGD) \cite{robbins1951stochastic} with a batch size of $128$, learning rate of $0.1$, weight decay of $0.0005$, and Nesterov momentum \cite{nesterov_method_1983} of $0.9$. We train ResNet-18 and ResNet-20 classifiers for $200$ epochs and multiply the learning rate by $0.2$ at epochs $\{60, 120, 160\}$. For SVHN, we use the same optimizer with a weight decay of $0.0001$. We train ResNet-18 classifiers for $120$ epochs and multiply the learning rate by $0.1$ at epochs $\{30, 60, 90\}$.

\paragraph{Data Augmentation.}  For CIFAR-10/100, we use standard data augmentation akin to~\cite{zagoruyko_wide_2016}. That is, we zero-pad images with $4$ pixels on each side, take a random crop of size $32 \times 32$, and then mirror the resulting image horizontally with ${50}\%$ probability. We use such data augmentation for both training the generative and the classifier models. Hence, our normalizing flows are capable of encoding-decoding operations on augmented samples as well. Advanced data augmentation baselines we use in Table \ref{tab:lowcifar10} \cite{devries2017cutout, zhang2018mixup}, also include the same standard data augmentations. However, the VAE-GAN based approach~\cite{stutz2019disentangling} does not use data augmentation in their generative model. To provide a more direct comparison between the performance of two generative models, in \S\ref{app:cifar10} we conduct an additional study without any data augmentations.

\subsection{Metrics}\label{app:metrics}

\paragraph{Fr\'echet Inception Distance.}\label{par:fid}
FID \cite{heusel2017gans} aims at comparing the synthetic samples $x \sim p_g$---where $p_g$ denotes the distribution of the samples of the given generative model, with those of the training data of $x \sim p_d$ in a feature space. The samples are embedded using the first several layers of the Inception network. Assuming $p_g$ and $p_d$ are multivariate normal distributions, it then estimates the means $\vm_g$ and $\vm_d$ and covariances $C_g$ and $C_d$, respectively for  $p_g$ and $p_d$ in that feature space. Finally, FID is computed as: 
\begin{align}\tag{FID}
\mathbb{D}_{\text{FID}}(p_d, p_g) &\approx d^2((\vm_d, C_d), (\vm_g, C_g )) =  \|\vm_d - \vm_g\|_2^2 + Tr(C_d + C_g - 2(C_dC_g)^{\frac{1}{2}}), 
\end{align}
where $d^2$ denotes the Fr\'echet Distance.
Note that as this metric is a distance, the lower it is, the better the performance.
We used the implementation of FID\footnote{\url{https://github.com/mseitzer/pytorch-fid}} in PyTorch.

\section{Additional Results}\label{app:results}

\subsection{Results on MNIST}
Table~\ref{tab:mnist_generalization} summarizes our results on MNIST in full data regime. Although the baseline has a very good performance on this dataset, we observe improved generalization.

\begin{table}[!htbp]
    \centering
        \begin{tabular}{lcccc}
    \toprule
    \textbf{Perturbation}                      &  \textbf{Train Accuracy} & \textbf{Train Loss} &  \textbf{Test Accuracy} &  \textbf{Test Loss}  \\
    \midrule
    Standard                                               &  $99.80$&$0.0069$    &  $99.24$&$0.0288$  \\
    Randomized-LA, $\ell{=}{\ell_\infty}, \epsilon{=}0.15$                                  &  $99.78$ & $0.0076$    &  $99.28$&$0.0262$  \\
    Adversarial-LA, $\ell{=}{\ell_\infty}, \epsilon{=}0.05, \alpha{=}0.01, k{=}10$     &  $99.26$&$0.0230$    &  $99.43$&$0.0216$   \\
    \bottomrule \\
    \end{tabular}
        \caption{Train and test accuracy ($\%$) as well as {loss} on \textbf{MNIST}. Comparison with standard training, versus our latent-space perturbations.
    }
    \label{tab:mnist_generalization}
\end{table}

\subsection{Additional Results on CIFAR-10}\label{app:cifar10}

\paragraph{Results without Data Augmentation.} To provide a direct comparison between two generative models and eliminate the effect of data augmentation, we run additional experiments. Table~\ref{tab:noaug_cifar10} shows results for our latent perturbations without any data augmentation to train the normalizing flow and the classifier. In line with our FashionMNIST results in \S\ref{subsec:stutz}, we observe that both randomized and adversarial latent attacks overperform the standard baseline and the VAE-GAN based approach. 

\begin{table}[h]\centering
\begin{tabular}{lc}
\toprule  
\textbf{Method}                                         & \textbf{Accuracy}  \\ \midrule
Standard                                         & $49.8$ \\
VAE-GAN            & $49.4$\\
Randomized-LA                             & $54.9$ \\
Adversarial-LA                            & $58.2$ \\
\bottomrule  \\
\end{tabular}
\caption{Test accuracy ($\%$) on \textbf{CIFAR-10}, in the \emph{low-data regime} ($5\%$ of training samples) without any data augmentation.}
\label{tab:noaug_cifar10}
\end{table}

\paragraph{Results with ResNet-20.} Table \ref{tab:resnet20} summarizes our results using the ResNet-20, on CIFAR-10. Inline with our ResNet-18 results in~\S\ref{subsec:cifar10}, we observe that both randomized and adversarial latent attacks overperform the standard baseline.

\begin{table}[h!]
\centering
\begin{tabular}{@{}lcc@{}}
\toprule  
\textbf{Method}                              & \textbf{Accuracy}  \\ \midrule
Standard                                     & $65.6$  & --\\
Randomized-LA, $\ell{=}\ell_2,\epsilon{=}25.$               & $72.7$  \\
Adversarial-LA, $\ell{=}\ell_2,\epsilon=.5$               & $77.1$  \\
\bottomrule \\
\end{tabular}
\caption{Test accuracy ($\%$) on \textbf{CIFAR-10} using ResNet-20, in the \emph{low-data regime} ($5\%$ of the training set).
}
\label{tab:resnet20}
\end{table}

\paragraph{Results with Different Attack Parameters.} In Table \ref{tab:cifar10_hyperparam}, we provide results with varying hyperparameters for the different attacks. Observe that for Adversarial-LA, in the \textit{high} perturbation setting---where $\epsilon=2.0$, the classifier still didn't fully fit to the training set, but performance in the test set is above the standard baseline. 

\paragraph{Multi-step Training.}
We run additional experiments where we sequentially apply different attack hyperparameters in multi-step training with weaker perturbations to increase the performance on the test set. The results are listed in Table \ref{tab:cifar10_hyperparam}, denoted with $+$.
\begin{table}[htbp]
    \centering
  \begin{tabular}{lcccc}
  \toprule
  \textbf{Perturbation}                      &  \textbf{Train Accuracy} & \textbf{Train Loss} &  \textbf{Test Accuracy} &  \textbf{Test Loss} \\
  \midrule
  \textit{Baselines:} &&\\[0.3ex] 
    Standard                                                 &  $100.0$ & $0.002$              &  $95.2$ & $0.194$   \\ [0.3ex]
    PGD, $\ell{=}{\ell_2}, \epsilon{=}2.0, \alpha{=}0.5, k{=}10$         &  $61.13$ & $0.895$              &  $75.7$ & $0.731$   \\ [0.3ex]
    PGD, $\ell{=}{\ell_\infty}, \epsilon{=}0.03, \alpha{=}0.008, k{=}10$ &  $77.3$ & $0.521$               &  $86.3$ & $0.442$  \\ [0.3ex]
  \midrule   \textit{Ours:} &&\\[0.3ex] 
  Randomized-LA, $\ell{=}{\ell_2}, \epsilon{=}10.0$                         &  $99.8$ & $0.007$               &  $95.8$ & $0.161$   \\ [0.3ex]
  Randomized-LA, $\ell{=}{\ell_\infty}, \epsilon{=}0.25$                    &  $99.5$ & $0.015$               &  $\mathbf{96.3}$ & $0.142$   \\ [0.3ex]
  \quad $+$Randomized-LA, $\ell{=}\ell_\infty, \epsilon{=}0.15$                           &  $100.0$ & $0.002$              &  $\mathbf{96.4}$ & $0.133$    \\ [0.6ex]
  \hdashline \vspace{-0.3cm}\\   Adversarial-LA, $\ell{=}{\ell_2}, \epsilon{=}1.0, \alpha{=}0.5, k{=}3$          &  $99.9$ & $0.005$               &  $\mathbf{96.6}$ & $0.126$  \\ [0.3ex]
  Adversarial-LA, $\ell{=}{\ell_2}, \epsilon{=}2.0, \alpha{=}1.5, k{=}2$          &  $89.1$ & $0.214$               &  $95.8$ & $0.134$  \\ [0.3ex]
  \quad $+$Adversarial-LA, $\ell{=}{\ell_2}, \epsilon{=}1.0, \alpha{=}0.75, k{=}2$  &  $99.2$ & $0.030$               &  $96.5$ & $0.114$  \\ [0.3ex]
  \quad $+$Adversarial-LA, $\ell{=}{\ell_2}, \epsilon{=}0.75, \alpha{=}0.5, k{=}2$  &  $99.7$ & $0.011$               &  $\mathbf{96.7}$ & $0.115$   \\ [0.3ex]
  \quad $+$Randomized-LA, $\ell{=}\ell_\infty,\epsilon{=}0.25$                           &  $100.0$ & $0.002$              &  $96.5$ & $0.132$   \\ [0.3ex]
  \quad \quad $+$Randomized-LA, $\ell{=}{\ell_2}, \epsilon{=}10.0$            &  $100.0$ & $0.002$              &  $96.6$ & $0.131$   \\ [0.3ex]
  \bottomrule \\
  \end{tabular}
  \caption{Train and test accuracy ($\%$) as well as {loss} on \textbf{CIFAR-10} using ResNet-18. All of the models are trained with the same hyperparameters listed in~\S\ref{app:hyperparams}. Perturbations listed with the $+$ sign indicates a multi-step training. For example, last row lists the result of the model trained with $P_{adv}^{\ell_2}, \epsilon=2.0, \alpha=1.5, k=2$ for 130 epochs, $P_{rand}^{\ell_\infty}, \epsilon=0.25$ for 40 epochs and $P_{rand}^{\ell_2}, \epsilon=10.0$ for 30 epochs. Note that, regardless of multi-step training, the hyperparameters, including the total number of training epochs ($=200$), remain fixed across the experiments.
    }
  \label{tab:cifar10_hyperparam}
\end{table}

\end{document}